# One-Shot Learning in Discriminative Neural Networks

**Jordan Burgess**
Department of Engineering
University of Cambridge
jb689@cam.ac.uk

**James Robert Lloyd**
Qlearsite
London, UK
james.lloyd@qlearsite.com

**Zoubin Ghahramani**
Department of Engineering
University of Cambridge
zoubin@eng.cam.ac.uk

## 1 Introduction & Related Work

We consider the task of one-shot learning of visual categories, or more generally, learning to classify images with few examples of particular classes. The currently dominant image classification paradigm of supervised deep learning performs well only when data is abundant. In this paper we explore a Bayesian procedure for updating a pretrained convnet to classify a novel image category for which data is limited. We demonstrate that the approach is competitive with state-of-the-art methods whilst also being consistent with 'normal' methods for training deep networks on large data.

Several approaches to one-shot learning have been noted as failing to beat a simple nearest-neighbour classifier [8]. Recent approaches of the problem have used relatively complicated architectures such as memory augmented neural networks [9, 10] or siamese networks [5]; or have been specialised for the task of one-shot learning [10].

Fei-Fei et al. [2] demonstrated one-shot learning as a Bayesian update to an image classification model with a prior based on categories learned with lots of data. Our work is an modern update of this work, applying this technique to deep convolutional networks.

## 2 Bayesian Updates of a Pretrained Convnet

We use a convnet, pretrained on a base dataset $\mathcal{D}_{\text{base}}$ (e.g. ImageNet [1]), as the basis for transferring knowledge to our new task. We decompose this convnet into a fixed feature extractor and a softmax classifier with weight matrix $W_{\text{base}}$.

For each new category, we need to specify $D + 1$ connections between the output neuron and the $D$-dimensional embedding plus a bias term. We learn these weights in a Bayesian fashion, using a prior based on the weights learned for other categories.

We model $W_{\text{base}}$ as a multivariate Gaussian, chosen to capture the correlation structure between the high level features, and set this as the prior on each new weightvector: $p(\mathbf{w}_i) = \mathcal{N}(\bar{\mu}, \bar{\Sigma} + \lambda \mathbb{I})$, where $\bar{\mu}$ and $\bar{\Sigma}$ are the sample mean and the sample covariance matrix of $W_{\text{base}}$ and $\lambda$ acts as a regulariser and ensures the covariance matrix is full rank.

To infer the posterior weights, we experimented with an MCMC-based sampling technique, NUTS [3], and a variational technique, ADVI [7].

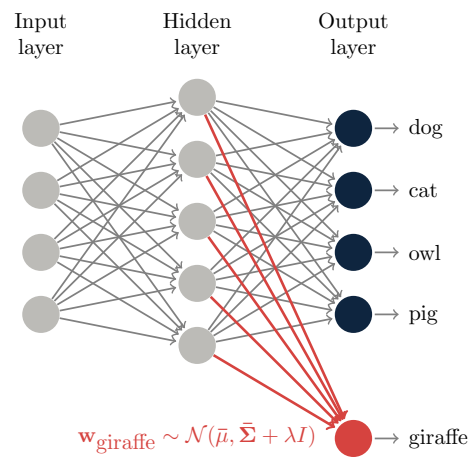

Figure 1: A pretrained convnet is used as a fixed feature extractor and the prior on the classifier weights is modelled on the existing weights.

## 3 Evaluation

We test the performance on a 5-way classification task, after having seen $n$ examples each category in training. The novel categories come from unseen examples of the same dataset, so as to reduce the dataset bias introduced. We use the tiny image dataset *CIFAR-100* [6] which we split 80:20 (in alphabetical order) into $\mathcal{D}_{\text{base}}$ and $\mathcal{D}_{\text{novel}}$ which contain examples of images $x_i$ and target labels $t_i$ from the category sets $C_{\text{base}}$ and $C_{\text{novel}}$[1]. The evaluation procedure is then:

1. sample 5 labels from $C_{\text{novel}}$
2. sample $n$ examples of each label to create small training data set $\mathcal{S} = \{(x_i, t_i) | i = 1, ..., 5n\}$
3. find posterior weights of each object category in light of $n$ observations
4. evaluate 5-way classification performance on all the remaining examples in $\mathcal{D}_{\text{novel}}$
5. repeat 20 times to establish performance bounds

### 3.1 Inference methods and sensitivity to the strength of prior

Using the faster but more approximate ADVI technique resulted in a $\approx 4\times$ speed up but a 2-percentage-point performance decrease compared to MCMC. A full-rank version of ADVI which does not approximate the posterior with diagonal Gaussians may be preferable for future work.

Overly broad priors ($\lambda \geq 10^0$) performed only marginally better than 'dumb priors', and overly precise priors ($\lambda \leq 10^{-4}$) degraded performance particularly after 20 training examples when the data has strong opinions about the weights. The results were otherwise fairly insensitive to $\lambda$ within this range, but the effect may be obscured by the large variance and so we're working on removing this free parameter.

### 3.2 Performance comparison

We compare our performance against alternative approaches for $n$-shot learning (see Figure 2a):

- '**Informed Prior**': our method of placing a multivariate Gaussian prior on the network weights of a softmax classifier.
- '**Dumb prior**': as above but fitting a spherical zero-centred Gaussian prior.
- '**Baseline softmax**': a standard softmax classifier initialised with Gaussian noise and optimised through gradient descent, with L2 regularisation of 0.001.
- '**Cosine**': Nearest neighbour classifier on the extracted image features by cosine distance.

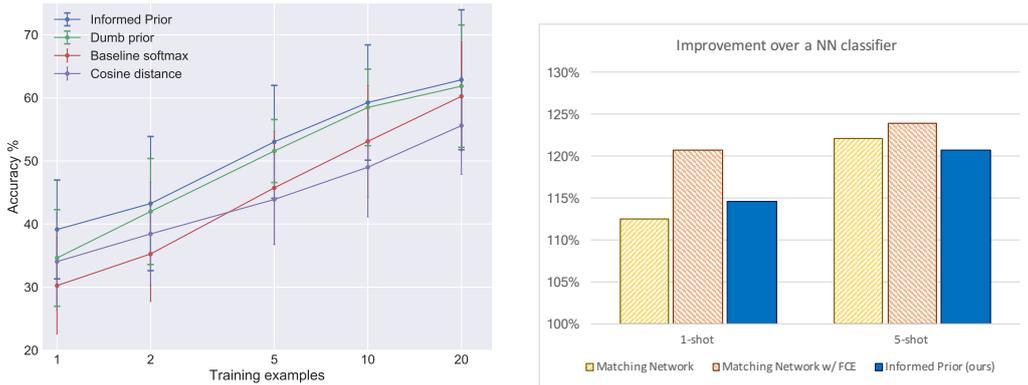

(a) Our informed prior gives significant improvement over just being Bayesian. Cosine distance is informationally inefficient with more data. (1 s.d. error bars)

(b) Indirect comparison against state-of-the-art one-shot learning technique of Matching Networks [10] using a nearest-neighbour classifier as a reference point for similar datasets.

Figure 2: 5-way classification results

---

[1]As there are no pretrained networks on 'CIFAR-80', we adapted and trained the 'quick-cifar10' model distributed with Caffe [4]. The model achieved 30.4% top-1 test accuracy.

A chance classifier would achieve 20% accuracy. A nearest neighbour classifier using cosine distance of the extracted features $\phi(x)$ achieves 34% after one shot. In contrast, training a softmax classifier by gradient descent, *with regularisation*, only manages 30%. Our one-shot learning method, achieves almost 40% accuracy on this challenging task.

### 3.3 Indirect comparison to the state-of-the-art

We can indirectly compare our method to the Matching Networks method proposed by Vinyals et al. [10]. Although the datasets differ in source[2], the training/test procedure is comparable. After one shot, a nearest-neighbour classifier with cosine distance metric achieves 36.6% on their *miniImageNet* dataset and 34.1% on our *CIFAR-100* dataset. The performance increase over this reference is shown in Figure 2b. Our 'informed prior' appears to be competitive with this state-of-the-art specialised model. However, part of their method is to allow the network to change the embedding of a test example as a function of the training examples with fully conditional embeddings (FCE). This improves their performance on the *miniImageNet* dataset shown, but it is noted that FCE does not seem "to help much" when tested on other datasets. We are working towards making this comparison exact.

## 4 Conclusions

One-shot learning continues to be a challenging problem for machine learning models. Non-parametric methods (combined with deep non-linear embeddings) often perform best in this low data regime. However, these methods become relatively weak if more data becomes available. Our approach, which is presented as a Bayesian learning procedure on a pretrained convnet, seemingly combines the best of both worlds. After a single data example, our 'informed' prior enables the classifier to perform similarly well to models highly specialised for this task. Yet, when more data is observed, the distribution over the weights collapses to a point mass and our model reduces to precisely the same model as a conventionally-trained deep convolutional network.

## References

[1] J. D. J. Deng, W. D. W. Dong, R. Socher, L.-J. L. L.-J. Li, K. L. K. Li, and L. F.-F. L. Fei-Fei. ImageNet: A large-scale hierarchical image database. *2009 IEEE Conference on Computer Vision and Pattern Recognition*, 2009.

[2] L. Fei-Fei, R. Fergus, and P. Perona. One-shot learning of object categories. *IEEE Transactions on Pattern Analysis and Machine Intelligence*, 2006.

[3] M. Hoffman and A. Gelman. The No-U-Turn Sampler: Adaptively Setting Path Lengths in Hamiltonian Monte Carlo. *Journal of Machine Learning Research*, 2014.

[4] Y. Jia, E. Shelhamer, J. Donahue, S. Karayev, J. Long, R. Girshick, S. Guadarrama, and T. Darrell. Caffe. In *Proceedings of the ACM International Conference on Multimedia*, 2014.

[5] G. Koch, R. Zemel, and R. Salakhutdinov. Siamese Neural Networks for One-shot Image Recognition. *ICML Deep Learning workshop*, 2015.

[6] A. Krizhevsky. Learning Multiple Layers of Features from Tiny Images. *Technical Report, Science Department, University of Toronto*, 2009. ISSN 1098-6596. doi: 10.1.1.222.9220.

[7] A. Kucukelbir, A. Gelman, and D. M. Blei. Automatic Differentiation Variational Inference. *arXiv preprint:1603.00788v1*, 2016.

[8] O. Rippel, M. Paluri, P. Dollar, and L. Bourdev. Metric Learning with Adaptive Density Discrimination. *International Conference on Learning Representations (ICLR)*, 2016.

[9] A. Santoro, S. Bartunov, M. Botvinick, D. Wierstra, and T. Lillicrap. One-shot Learning with Memory-Augmented Neural Networks. *arXiv preprint arXiv:1605.06065*, 2016.

[10] O. Vinyals, C. Blundell, T. Lillicrap, K. Kavukcuoglu, and D. Wierstra. Matching Networks for One Shot Learning. *arXiv preprint arXiv:1606.04080*, 2016.

---

[2]They derive a new dataset, *mini*ImageNet which consists of 100 classes, 600 examples each (like CIFAR-100), of 84 x 84 pixel images (larger than CIFAR-100) which they split 80:20 into base/novel classes (as we do).